\documentclass{article}

\usepackage[preprint]{corl_2026}

\usepackage{graphicx}
\usepackage{float}
\usepackage{amsmath}
\usepackage{amssymb}
\usepackage{mathtools}
\usepackage{amsthm}
\usepackage{multirow}
\usepackage{booktabs}
\usepackage{adjustbox}
\usepackage{siunitx}
\usepackage{wrapfig}
\usepackage{makecell}

\newcommand{\ours}{{OASIS}}

\title{OASIS: From Simulation Data Collection to Real-World Humanoid Loco-Manipulation}

\author{
  Zehao Yu\textsuperscript{1,2} \quad
  Jiakun Zheng\textsuperscript{1,3} \quad
  Weiji Xie\textsuperscript{1,4} \\
  \vspace{5pt}
  \textbf{Jiyuan Shi}\textsuperscript{1}\quad
  \textbf{Chenyun Zhang}\textsuperscript{1}\quad
  \textbf{Chenjia Bai}\textsuperscript{1\dag{}}\quad 
  \textbf{Xuelong Li}\textsuperscript{1\dag{}}\\
  \textsuperscript{1}Institute of Artificial Intelligence (TeleAI), China Telecom \quad
  \textsuperscript{2}Fudan University \\
  \textsuperscript{3}East China University of Science and Technology\quad 
  \textsuperscript{4}Shanghai Jiao Tong University \\
  \textsuperscript{\dag}Corresponding author\\
}

\begin{document}
\maketitle

\vspace{-2em}
\begin{figure}[ht]
  \centering
  \includegraphics[width=0.95 \linewidth]{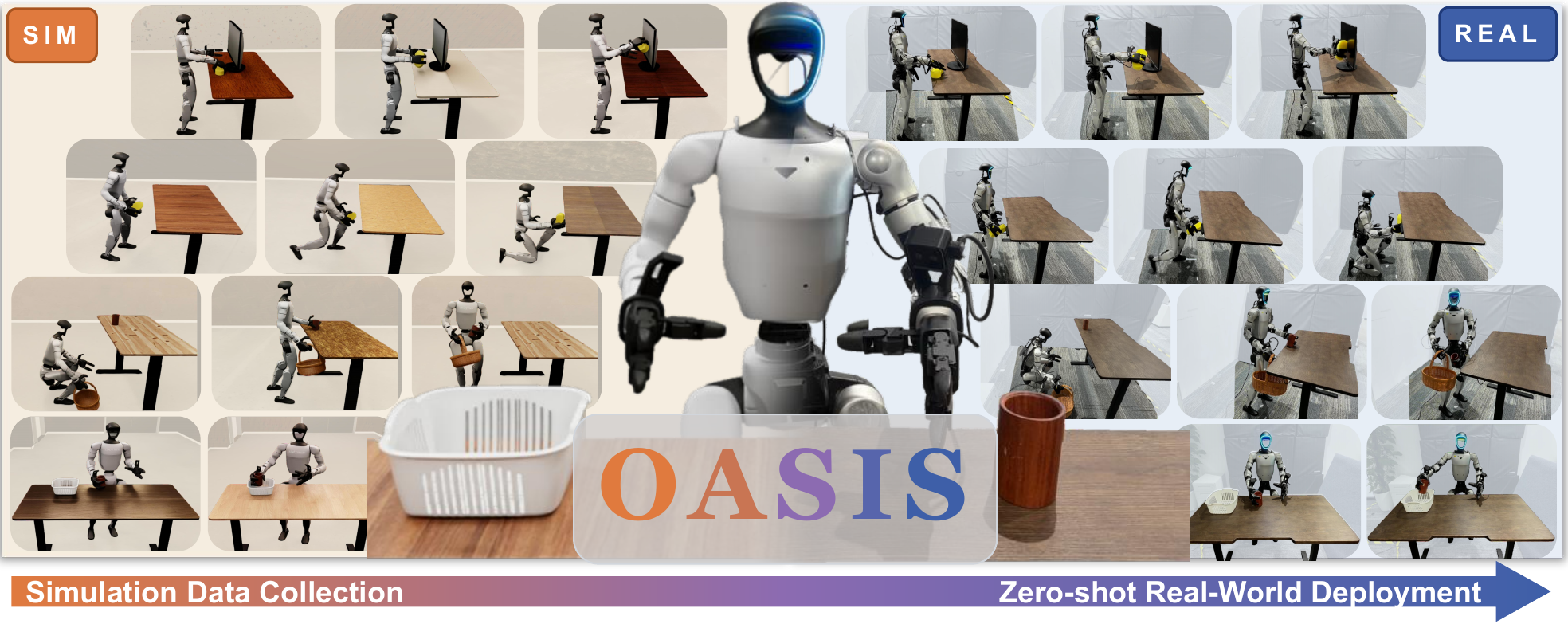}
  \caption{OASIS collects whole-body demonstrations entirely in simulation and deploys the visuomotor policy zero-shot on the real Unitree G1 humanoid across diverse loco-manipulation tasks.}
  \label{fig:cover}
\end{figure}

\begin{abstract}
Recent progress in robot manipulation has been largely driven by learning from large-scale demonstrations. For humanoid robot loco-manipulation tasks, however, existing data sources force an unsatisfying tradeoff between trajectory quality and scalability. Real-world teleoperation provides the highest-quality trajectories but requires dedicated physical space and time-consuming scene resets. Simulation offers an alternative way out of this dilemma: it can produce clean, embodiment-aligned data at scale without any physical hardware. In this paper, we propose \ours{}, a simulation-data-driven framework for humanoid loco-manipulation. \ours{} automatically reconstructs realistic object assets from real-world images using a 3D generative model. Based on these assets, trajectories are first collected through teleoperation in simulation, and then augmented under diverse domain randomizations in a post-processing stage. With the resulting simulation data, we further design a hierarchical visuomotor policy for humanoid loco-manipulation. Extensive experiments on the real humanoid robot show that, under zero-shot deployment, the policy trained on our simulation data achieves higher success rates on most tasks than that trained on real-robot teleoperation data, owing largely to the broad lighting and environmental variations covered by our simulation rendering, which real-robot data fails to capture.
\end{abstract}



\section{Introduction}
Humanoid robots are expected to take on a wide range of tasks in everyday human environments~\cite{gu2026humanoid}, where locomotion and manipulation must be tightly coordinated to act effectively~\cite{fu2024humanplus, he2024omnih2o}. However, robust and generalizable loco-manipulation ultimately depends on large-scale, high-quality demonstration data, which current humanoid platforms still largely lack~\cite{o2024openx, brohan2022rt1, zitkovich2023rt2}.

\begin{wrapfigure}{r}{0.5\linewidth}
  \centering
  \vspace{-10pt}
  \includegraphics[width=\linewidth]{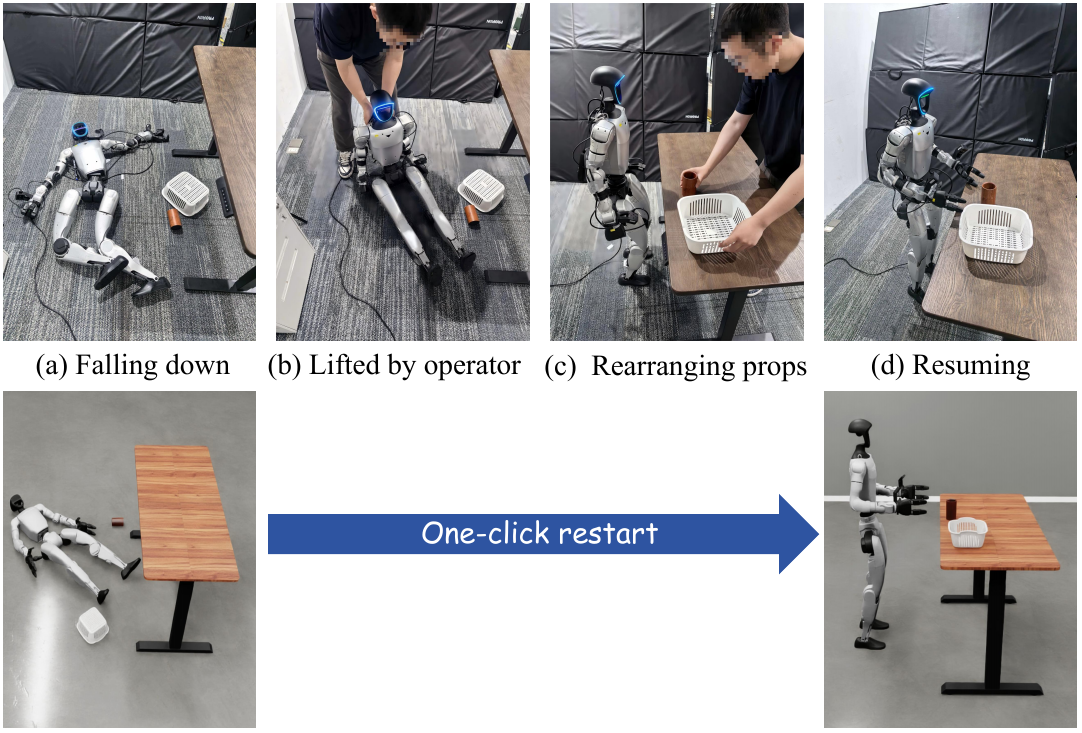}
  \caption{\textbf{Failure recovery: real robot vs. simulation.} Recovering from a failure on a real humanoid requires a tedious manual sequence — (a) falling down, (b) lifted by the operator, (c) rearranging props, and (d) resuming. In contrast, simulation supports one-click restart, restoring the scene instantaneously.}
  \label{fig:recovery}
  \vspace{-10pt}
\end{wrapfigure}

To obtain the demonstration data required for humanoid manipulation, prior work has explored a range of sources, including human videos~\cite{nair2022r3m, wang2023mimicplay}, egocentric recordings~\cite{grauman2022ego4d, kareer2025egomimic, zheng2026egoscale, hoque2025egodex}, and real-robot teleoperation~\cite{shi2026egohumanoid, bai2026hex, bjorck2025gr00t, ding2025humanoidvla, wei2026psi_0}. Among these, real-robot teleoperation has been the most widely used~\cite{ze2025twist,ze2025twist2,luo2025sonic, li2026omniclone}, as the operator directly drives the robot to complete the task, yielding trajectories that are precisely aligned with the robot's embodiment and inherently accompanied by action supervision. However, collecting teleoperation data on real robots is time-consuming, resource-intensive, and hard to scale. First, large-scale collection requires a substantial number of expensive robots and supporting equipment, along with correspondingly large physical spaces, resulting in high financial and spatial overhead. Second, physical interaction itself makes the collection process fragile and inefficient. In long-horizon tasks, any failure or need for repositioning requires manual reconfiguration of both the robot and the environment, significantly slowing collection, whereas simulation~\cite{makoviychuk2021isaacgym, NVIDIA_Isaac_Sim, todorov2012mujoco} allows instantaneous reset, as shown in Fig.~\ref{fig:recovery}. Moreover, operational errors during physical interaction often damage robot hardware or objects in the scene. 

To address these limitations, we introduce \ours{}, a framework that learns humanoid loco-manipulation policies from data collected entirely in simulation. Given reference images of real objects, it synthesizes 3D meshes with a 3D generative model and estimates their physical dimensions and material properties with a vision-language model (VLM). The resulting assets closely match their real counterparts, enabling diverse and physically plausible simulation scenes to be built at scale. Built on these assets, \ours{} adopts a two-stage decoupled design. In the first stage, the operator teleoperates a humanoid robot in simulation in real time from a first-person view through VR devices such as PICO 4U~\cite{pico4ultra_2023}, a portable virtual reality system that captures the operator's full-body pose through a headset, a pair of handheld controllers, and two ankle-mounted trackers, obviating the need for dedicated motion-capture studios. To preserve real-time responsiveness, the VR headset receives a lightweight rendering for operator feedback, while only the state sequences of the robot and the objects in the scene are recorded. In the second stage, the recorded states are replayed offline and rendered at high fidelity for training. Textures, lighting, and camera extrinsics are randomized in the process, turning each teleoperated trajectory into a diverse set of visually distinct training samples. This decoupling separates the cost of teleoperation from the size of the resulting dataset, so a small amount of operator time produces a large and visually diverse training set.

We build a hierarchical visuomotor policy based on our system. The high-level planner is a Flow Matching policy that predicts reference motion commands from visual observations, and the low-level controller converts these commands into target joint angles. We validate our system on a real humanoid robot. The high-level planner is trained purely on simulation data and successfully accomplishes several tasks zero-shot. It also demonstrates adaptability to real-world perturbations such as camera motion blur and background clutter.
    
The main contributions of this paper are as follows. First, we present \ours{}, a humanoid loco-manipulation framework that implements a novel pipeline in which control policies are learned entirely from teleoperation data collected in simulation. Second, we design a scalable data collection system that enables efficient demonstration collection in simulation. Third, through real-robot experiments, we demonstrate that data collected with \ours{} enables effective zero-shot transfer to real robots on multiple tasks.

\section{Related Work}
\label{sec:related_work}

\subsection{Humanoid Loco-Manipulation}

Humanoid loco-manipulation requires coordinated locomotion, whole-body manipulation, and task-level reasoning, and remains challenging from both the execution and the data-supervision sides. To shoulder the engineering burden, recent works standardize humanoid policy learning into reproducible workflows~\cite{zhao2026agile}. In parallel, a growing body of work builds generalist humanoid vision-language-action (VLA) policies trained on heterogeneous mixtures of human videos, synthetic data and teleoperated trajectories, typically pairing a vision-language backbone with a fast action expert and a dedicated whole-body tracking controller~\cite{bjorck2025gr00t, ding2025humanoidvla, wei2026psi_0, fu2026demohlm}. More recent efforts further refine this paradigm by introducing humanoid-aligned state representations for cross-embodiment learning~\cite{bai2026hex} or unified latent VLAs for manipulation-aware locomotion~\cite{jiang2025wholebodyvla}. Alongside this, to bypass the bottleneck of robot teleoperation, another line of work collects robot-free demonstrations through portable rigs, wearable exoskeletons, or egocentric capture devices, and bridges the human-humanoid embodiment gap via view and action alignment~\cite{nai2026humi, zhong2025humanoidexo, shi2026egohumanoid}; even teleoperation-based systems have shifted toward portable, mocap-free setups to make whole-body data collection scalable~\cite{ze2025twist2}. Despite this progress, both robot-centric teleoperation and robot-free human capture remain time-consuming and physically expensive. In contrast, we treat high-fidelity simulation as a scalable source of whole-body loco-manipulation data: by automatically constructing physically plausible scenes from generative 3D assets and decoupling trajectory collection from photorealistic rendering, each teleoperated trajectory is expanded into a large number of visually diverse training samples, on which we train a Flow Matching policy.

\subsection{Simulation Data Collection For Robot Learning}

The high cost of real-robot data collection has motivated growing efforts to use simulation as a scalable training source~\cite{nasiriany2024robocasa, wang2024gensim}. One line of work automates the construction of simulated tasks and assets via foundation models~\cite{wang2024robogen}. Another scales data through trajectory augmentation, replaying a few human demonstrations into many new initial conditions. MimicGen~\cite{mandlekar2023mimicgen} establishes this paradigm for tabletop manipulation, and DexMimicGen~\cite{jiang2025dexmimicgen} extends it to bimanual dexterous setups. However, both remain restricted to fixed-base or upper-body settings. Recent humanoid-specific efforts further explore simulation data for whole-body policies, yet each carries its own bottleneck. GR00T N1~\cite{bjorck2025gr00t} augments its corpus with synthetic data, but the simulated portion is dominated by simple bimanual tabletop tasks. VIRAL~\cite{he2025viral} collects data via reinforcement learning in simulation, but the RL acquisition process itself is expensive, requiring carefully shaped rewards and long training per behavior. 
In contrast, we automatically construct physically plausible scenes from generative 3D assets, collect whole-body trajectories via simulation teleoperation, and successfully train policies that transfer to real-robot.
	
\begin{figure}[ht]
  \centering
  \includegraphics[width=0.95 \linewidth]{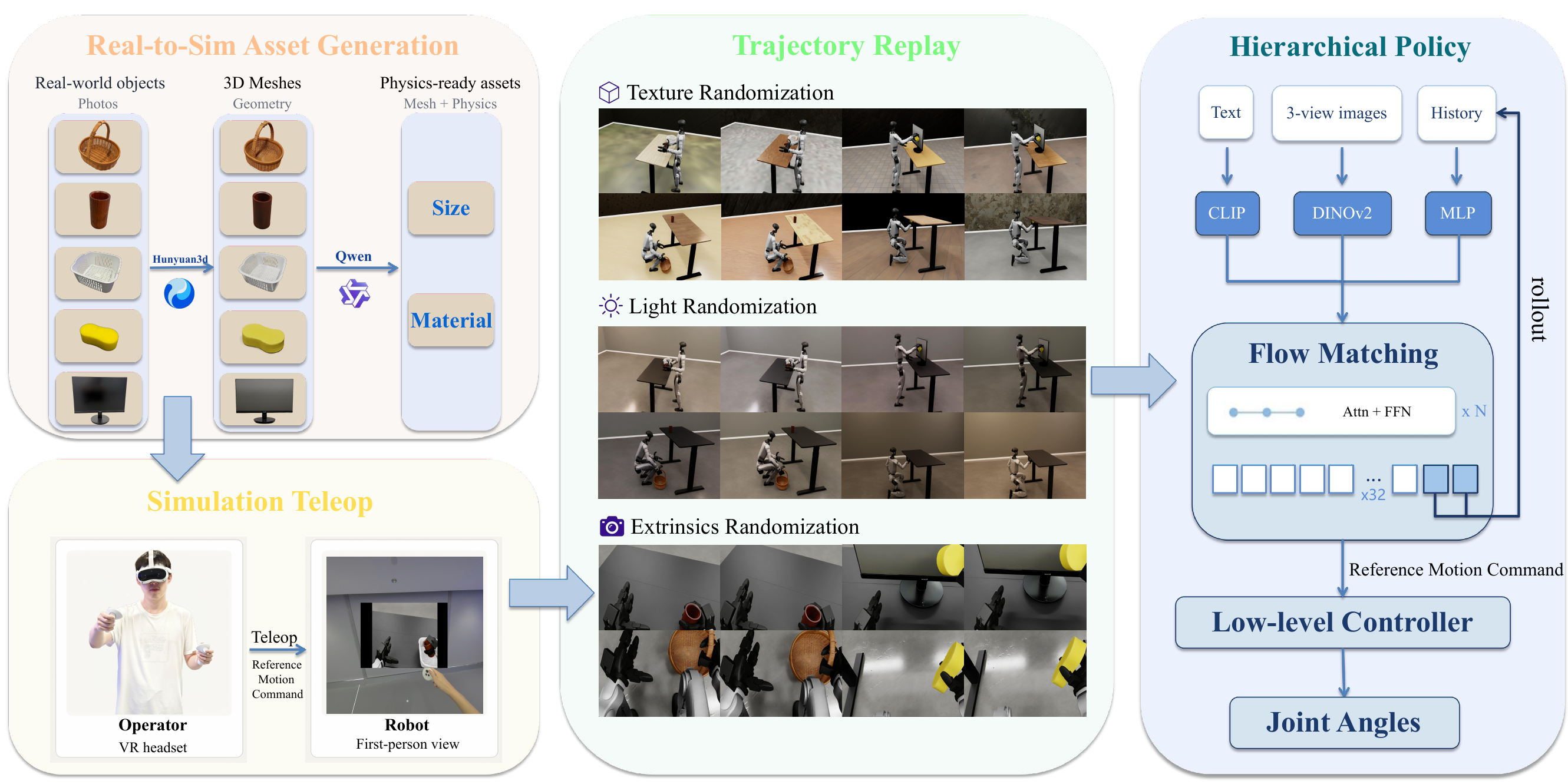}
  \caption{\textbf{Overview of \ours{}.} Our framework consists of four stages. First, we reconstruct physics-ready simulation assets from single-view photos of real objects. Second, demonstration trajectories are collected in simulation via VR teleoperation. Third, these trajectories are replayed with texture, lighting, and camera-extrinsics randomization for visual augmentation. Finally, a hierarchical policy is trained on the augmented data, where a high-level Flow Matching predicts reference motion command from multimodal observations, and a low-level controller tracks them as joint angles in a closed loop.}
  \label{fig:method}
\vspace{-1.5em}
\end{figure}

\section{Method}
\label{sec:method}

\subsection{Overview}
OASIS is a simulation-data-driven framework for humanoid loco-manipulation, consisting of automated simulation scene construction, a two-stage teleoperation-and-rendering data collection pipeline, and a hierarchical whole-body policy trained on the resulting data for zero-shot real-robot deployment.
In this section, we detail (i) how OASIS collects scalable simulation data, and (ii) how we learn a hierarchical whole-body policy that transfers to the real robot.

\subsection{Data Collection}

\subsubsection{Simulation Scene Construction}
\label{sec:simulation_scene_construction}
Moving data collection into simulation removes the dependence on physical hardware, but it introduces a new bottleneck: every task requires a corresponding simulation scene with realistic, physically plausible objects, and constructing such scenes by hand is itself labor-intensive. To eliminate this bottleneck, we build an automated asset generation pipeline.

\textbf{Real-to-Sim Asset Generation.} Given reference images of real-world objects, 
we first leverage Hunyuan3D~\cite{zhao2025hunyuan3d}, an advanced large-scale 3D synthesis system for generating high-resolution textured 3D assets. The outputs of the generative model consist solely of meshes and texture maps, lacking both physical scale and material properties. 
To recover these attributes, we further leverage the strong prior knowledge of Qwen3-VL~\cite{bai2025qwen3}, a vision-language model with strong visual reasoning capabilities over object geometry, materials, and physical properties. Given the reference image and a category description of the object, it is prompted with a structured template to produce reasonably accurate estimates of the object's physical dimensions and material category.

\textbf{Physics Parameter Assignment.}
The predicted dimensions rescale the normalized mesh to its physical size. Meanwhile, the material category serves as an index into a predefined table to retrieve the effective density, friction, and restitution coefficients, as detailed in Appendix A. From these, mass and inertia are computed under a uniform-density assumption, while friction and restitution are attached to the collision body. To account for estimation errors, all physical properties are randomized around their predicted values during data collection.

\subsubsection{Teleop Trajectory Collection}
With the simulation scene built from the generated assets, we collect humanoid manipulation trajectories through VR-based teleoperation. Human operators control the simulated humanoid via VR devices such as PICO 4U, while the robot's head-camera stream is transmitted to the headset in real time as a first-person view.

The operator's motions are retargeted to the humanoid by GMR~\cite{GMR} to produce reference whole-body motions, which are then input to Teleopit~\cite{botrunner64_teleopit_2025}, an open-source reinforcement learning-based whole-body controller, to drive the simulated humanoid to execute the corresponding actions. To maintain low-latency teleoperation, this stage employs the Real-Time rendering mode of IsaacSim~\cite{NVIDIA_Isaac_Sim}, which substantially reduces the rendering overhead while preserving sufficient visual fidelity, allowing the simulator to run at a high frame rate.

During data collection, two categories of data are recorded. The first is the whole-body kinematic state of the robot, together with the kinematic states of all interactive rigid bodies in the scene. These states are used to replay the trajectory in the second stage. The second is the reference motions retargeted by GMR, which are used to train the high-level policy.

\subsubsection{Scalable Trajectory Rendering}
We then collect diverse image observations paired with these recorded trajectories to construct the training dataset. 
Each trajectory is replayed offline and rendered under randomized visual conditions, expanding a single demonstration into a large number of visually diverse samples.
Free from the real-time constraint of teleoperation, the offline setting enables Path-Tracing rendering mode in IsaacSim, which produces higher-fidelity images. 
Specifically, we randomize background textures, the intensity and color temperature of environmental lighting, and the extrinsic parameters of the cameras. 

\subsection{Whole-Body Policy Learning}

\subsubsection{Model Architecture}
\label{sec:model_architecture}
Following TextOp~\cite{xie2026textop}, we represent the per-frame reference motion command $m_t \in \mathbb{R}^{67}$ as:
\begin{equation}
m_t =
\Big[
\phi(r_t),\;
\Delta \psi_t,\;
\Delta p_t^{\text{local}},\;
h_t,\;
q_t,\;
\Delta q_t
\Big],
\end{equation}
where $\phi(r_t) = 
[\sin(\text{roll}_t), \cos(\text{roll}_t)-1, \sin(\text{pitch}_t), \cos(\text{pitch}_t)-1]$ represents the trigonometric encoding of roll and pitch, $\Delta \psi_t = \text{yaw}_{t+1} - \text{yaw}_t$ denotes the per-frame yaw difference, $\Delta p_t^{\text{local}} = R_z(\text{yaw}_t)^\top (p_{t+1} - p_t)$ is the root translation in the local frame, $h_t$ represents root height, $q_t \in \mathbb{R}^{29}$ represents joint positions, and $\Delta q_t = q_{t+1} - q_t$ represents joint increments.

As illustrated in Fig.~\ref{fig:method}, our high-level planner is a Transformer-based, action-chunking policy that generates future motion sequences with Flow Matching~\cite{lipman2022flow}, and is coupled with a low-level controller in a hierarchical design. The denoiser takes three inputs: the text instruction, encoded by a frozen CLIP~\cite{radford2021learning} text encoder; three-view images, encoded by a frozen DINOv2~\cite{oquab2023dinov2} visual encoder; and robot proprioception over the most recent $H = 2$ frames, encoded by an MLP. These features are concatenated into a condition token sequence $c$, on which the denoiser predicts the whole-body reference motion $\mathbf{m}_{t:t+F} \in \mathbb{R}^{F \times 67}$ over the next $F = 32$ frames.

We train the denoiser $v_\theta$ with the Flow Matching objective, which regresses the constant-velocity field along the linear path between a Gaussian prior $\mathbf{a}_1 \sim \mathcal{N}(0, I)$ and the target action chunk $\mathbf{a}_0$:
\begin{equation}
\mathcal{L}_{\text{FM}}(\theta) =
\mathbb{E}_{\tau,\, \mathbf{a}_0,\, \mathbf{a}_1}
\Big[
\big\|\, v_\theta(\mathbf{a}_\tau, \tau, c) - (\mathbf{a}_1 - \mathbf{a}_0) \,\big\|_2^2
\Big],
\end{equation}
where $\mathbf{a}_\tau = (1-\tau)\,\mathbf{a}_0 + \tau\,\mathbf{a}_1$ and $\tau \sim \mathcal{U}(0,1)$. At inference, we generate actions by integrating the learned velocity field with an Euler solver using $10$ denoising steps. Consistent with teleoperation, the low-level controller Teleopit converts the reference motion into 29-DoF body joint angles; together with the 14-DoF hand joints, the system outputs 43-DoF whole-body joint angles.

\subsubsection{Training Recipe}
\label{sec:training_recipe}
\textbf{Reference Motion Commands as Proprioception.}
For the proprioception input, we use the reference motion commands rather than the robot state. The robot state reflects the trajectory already executed by the low-level controller, which inevitably carries tracking errors and noise; conditioning the planner on such signals lets these errors accumulate and feed back into planning. The reference commands, in contrast, provide a consistent and noise-free history, keeping the planner's input distribution identical between simulation and deployment.

\textbf{Curriculum-based Rollout Training.}
Since the planner predicts $F$ frames in a single pass, training only on ground-truth history leaves it unable to cope with the accumulated errors of its own predictions at inference, causing instability over long horizons. We therefore adopt a curriculum-based rollout mechanism: at each training step we sample $P = 4$ consecutive segments from the same sequence, where the first segment uses ground-truth history and each subsequent segment reuses its predecessor's last $H$ predicted frames with probability $p_{\text{rollout}}$. This probability stays at 0 for the first 20\% of training, letting the model first fit the conditional distribution on clean history, then increases linearly to 0.8. By exposing the model to its own prediction errors during training, this mechanism maintains stability under long-horizon autoregressive rollout at deployment.

\subsection{Deployment}
We deploy our system on a 29-DoF Unitree G1 humanoid, equipped with 7-DoF three-fingered dexterous hands. In addition to a Realsense D435i camera on the head, each wrist is fitted with an additional Realsense D405 camera. The high-level planner operates at 25 Hz on an NVIDIA RTX 4090 GPU, while the low-level controller executes the predicted 32-step action chunk at 50 Hz.

\begin{figure}[t]
  \centering
  \includegraphics[width=0.85 \linewidth]{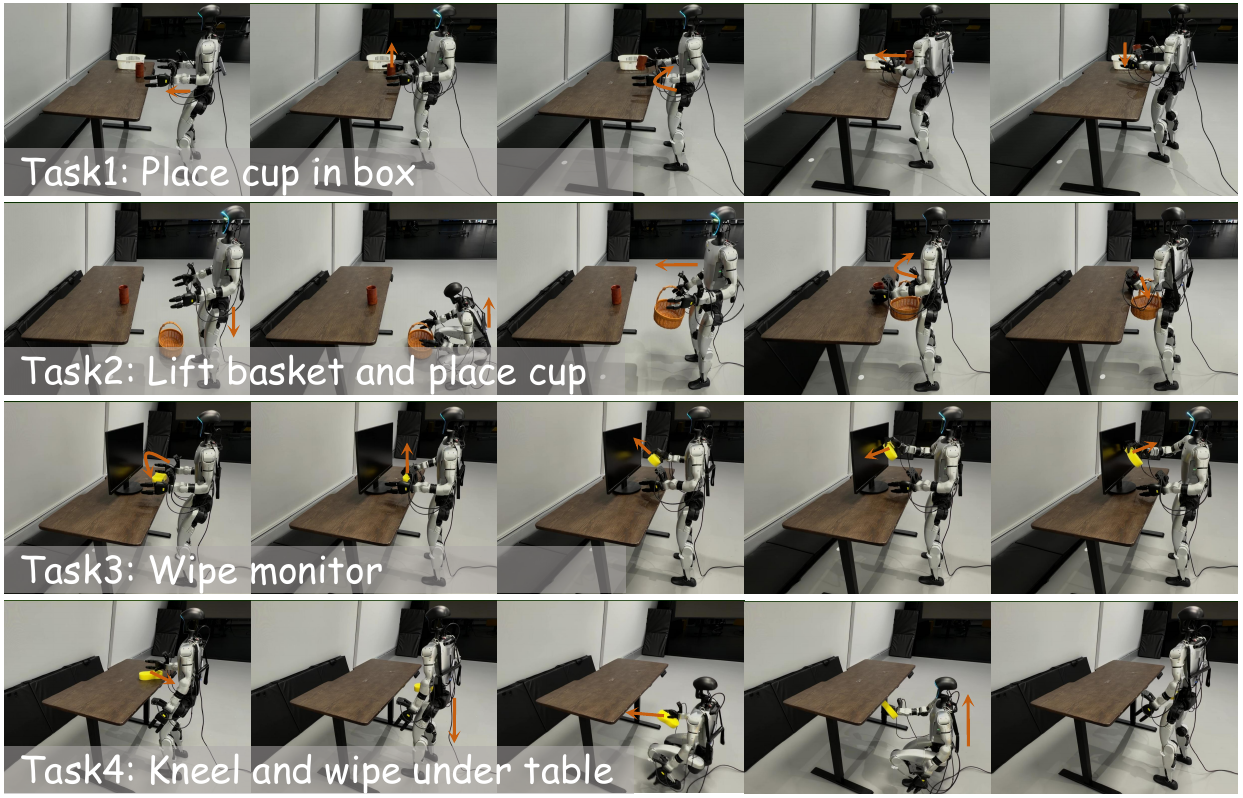}
  \caption{\textbf{Real-robot experiments on loco-manipulation tasks across different difficulty levels.}}
  \label{fig:real_robot_experiment}
\end{figure}

\section{Experiments}
\label{sec:experiments}
In this section, we conduct experiments on the Unitree G1 humanoid to answer the following questions:
\textbf{Q1:} Can \ours{} achieve higher data collection efficiency than real-robot teleoperation?
\textbf{Q2:} How does each component of the \ours{} data augmentation stage affect sim-to-real transfer?
\textbf{Q3:} How effective is simulation data from OASIS compared with real-robot data for humanoid loco-manipulation?

\subsection{Data Collection Efficiency}

\begin{table}[hb]
\centering
\begin{tabular}{lccc}
\toprule
Task & \ours{} (min) & Real (min) & Speedup \\
\midrule
Place Cup in Box                & 15.2 & 17.5 & 1.15$\times$ \\
Wipe Monitor                    & 19.1 & 26.8 & 1.40$\times$ \\
Lift Basket and Place Cup       & 25.2 & 40.2 & 1.60$\times$ \\
Kneel and Wipe Under Table      & 28.4 & 44.8 & 1.84$\times$ \\
\bottomrule
\end{tabular}
\vspace{4pt}
\caption{\textbf{Time taken to collect 50 successful trajectories per task with \ours{} versus real-robot teleoperation.} \ours{} is faster on every task, and the gap is larger on harder ones.}
\label{tab:collection_time}
\end{table}

To address \textbf{Q1}, we measure data collection efficiency in both simulation and the real world. Specifically, we use the same low-level controller and the same operator, and collect the same number of successful trajectories on the same tasks. As shown in Table~\ref{tab:collection_time}, collecting data with \ours{} is significantly faster than real-robot collection across all tasks, and the speedup grows with task difficulty.

Since both settings drive the same humanoid through the same interface, the time spent executing each task is comparable; the efficiency gap arises almost entirely from the overhead beyond each trajectory, which is unavoidable in the real world but nearly zero in simulation. In real-robot collection, after each attempt the operator must enter the workspace and reset every object to its initial configuration before the next trajectory can begin, and this overhead grows with the number of objects and the task length. In simulation, resets are instantaneous and fully automatic.

Moreover, physical interaction is fragile, and this fragility extends even to the manipulated objects. In our screen-wiping task, the robot makes frequent contact with a fragile monitor, and during real-robot collection any deviation in force or timing risks damaging it—in fact, we damaged a monitor due to excessive contact force, forcing the operator to proceed slowly and cautiously. In simulation, none of this is a concern: a damaged object can simply be reset, so the operator does not need to hold back.

\subsection{Ablations on Data Augmentation}
\ours{} augments each trajectory by applying vision randomization and rendering it multiple times, expanding a single demonstration into a large set of visually diverse training samples. To address \textbf{Q2}, we examine this component from two angles: the contribution of each randomization factor, and the number of renderings needed per trajectory. 

For the randomization factors, we compare disabling all randomization (w/o All), removing one factor at a time (texture, lighting, or camera extrinsics), and the full configuration (Ours). As shown in Table~\ref{tab:augmentation_ablation}, disabling all randomization causes the policy to almost completely fail to transfer, confirming that randomization is indispensable. Among the individual components, lighting contributes the most, since illumination differences are among the largest sim-to-real visual gaps. Importantly, the full combination outperforms every ablated variant, indicating that these randomizations target complementary aspects of the sim-to-real gap and are most effective when applied jointly.

The success rate rises steadily with more renderings and approaches saturation around 15--20, beyond which the gains taper off. We therefore render each trajectory into 20 environments to balance performance and overhead.

\begin{table}[ht]
\centering
\setlength{\tabcolsep}{4pt}
\small
\begin{tabular}{l*{8}{>{\centering\arraybackslash}m{0.85cm}}}
\toprule
& \multicolumn{4}{c}{Domain Randomization} & \multicolumn{3}{c}{Rendered Envs. per Traj.} & \\
\cmidrule(lr){2-5}\cmidrule(lr){6-8}
Task & w/o All & w/o Tex. & w/o Light & w/o Cam. & 5 & 10 & 15 & Ours \\
\midrule
Place Cup in Box & 0/10 & 5/10 & 3/10 & 7/10 & 4/10 & 5/10 & 8/10 & \textbf{8/10} \\
Lift Basket and Place Cup & 0/10 & 3/10 & 1/10 & 5/10 & 2/10 & 4/10 & 5/10 & \textbf{7/10} \\
Wipe Monitor & 1/10 & 5/10 & 4/10 & 7/10 & 5/10 & 7/10 & 7/10 & \textbf{8/10} \\
Kneel and Wipe Under Table & 1/10 & 4/10 & 4/10 & 6/10 & 4/10 & 7/10 & 10/10 & \textbf{10/10} \\
\midrule
Average Success Rate& 0.05 & 0.43 & 0.30 & 0.63 & 0.38 & 0.58 & 0.75 & \textbf{0.83} \\
\bottomrule
\end{tabular}
\vspace{4pt}
\caption{\textbf{Ablations on the data-augmentation stage.} All numbers are real-robot zero-shot success rates over 10 trials. The \emph{Ours} column denotes our final configuration, which applies all randomization and renders each trajectory under 20 randomized environments.}
\label{tab:augmentation_ablation}
\vspace{-2em}
\end{table}

\subsection{Effectiveness Of Simulation Data}
To address \textbf{Q3}, we evaluate policies on the real Unitree G1 across the loco-manipulation tasks shown in Fig.~\ref{fig:real_robot_experiment}, which span tabletop manipulation, whole-body lifting, and kneeling under-table wiping. For each task, we compare three sources of training data under the same total number of trajectories: simulation data from \ours{}, real-robot only, and an equal mixture of both.

\begin{wrapfigure}{r}{0.5\linewidth}
  \centering
  \vspace{-10pt}
  \includegraphics[width=\linewidth]{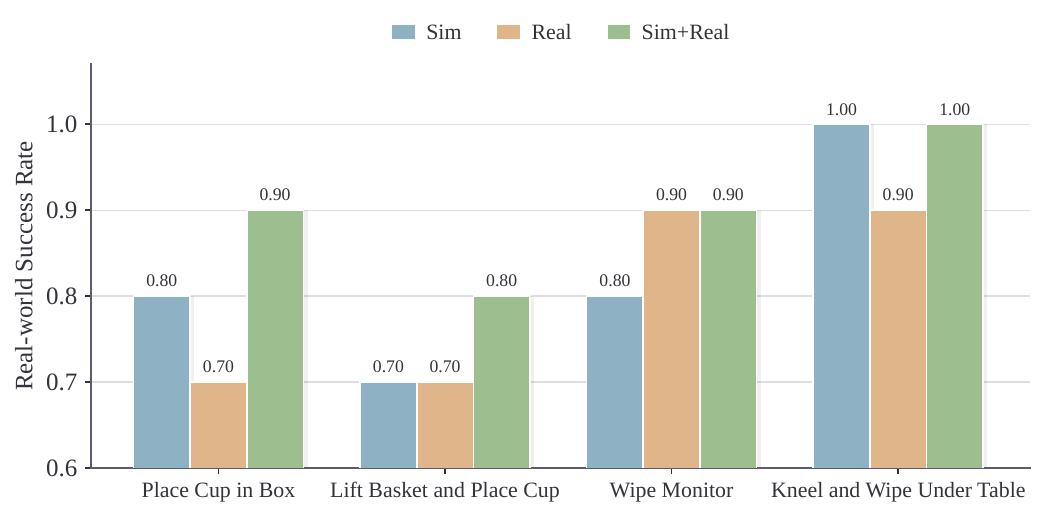}
  \caption{Real-world zero-shot success rates of policies trained on simulation data from \ours{}, real-robot data, and their equal mixture, using the same total of 50 trajectories per setting.}
  \label{fig:data_source}
  \vspace{-10pt}
\end{wrapfigure}

As shown in Fig.~\ref{fig:data_source}, the policy trained on simulation data alone achieves a real-robot success rate comparable to, and on some tasks higher than, the one trained on real-robot data. Since both use the same number of trajectories, this shows that the simulation data collected by \ours{} rivals real-robot data in supervision quality and can serve as an effective substitute, without the high time and hardware costs of real-robot collection. We attribute the cases where simulation even surpasses real data to visual diversity: real-robot data is collected in a relatively fixed environment, so the policy struggles once deployment conditions deviate from collection time, whereas the large-scale randomized re-rendering in simulation covers far richer visual conditions and yields stronger robustness.

Moreover, mixing the two sources under the same trajectory budget outperforms either alone. As the total data is unchanged, this gain stems not from more data but from their complementarity: simulation contributes large-scale, visually diverse samples for generalization, while real-robot data supplies the real interaction and perception characteristics that simulation cannot fully capture. Overall, simulation data alone supports high-performance real-robot deployment and further improves performance when combined with real data, highlighting the value of \ours{} as a scalable data source.

\section{Conclusion}
\label{sec:conclusion}
OASIS grounds simulated scenes in 3D-generated assets recovered from real-world images, and separates VR-based teleoperation from offline photorealistic rendering, so that each demonstration is expanded into a large set of visually diverse training samples without additional operator effort. On the Unitree G1 humanoid, data collection with OASIS runs up to 1.84$\times$ faster than real-robot teleoperation. Policies trained entirely on OASIS-generated data transfer zero-shot to the real robot, matching or surpassing those trained on real-robot data under the same trajectory budget. These results suggest that high-fidelity simulation, when paired with realistic asset generation and large-scale visual randomization, can serve as a practical and scalable alternative to real-robot teleoperation for humanoid loco-manipulation.

\section{Limitations}
\label{sec:Limitations}

While OASIS transfers from simulation to the real world zero-shot on loco-manipulation tasks, several limitations remain.

First, our augmentation only randomizes visual appearance and leaves trajectories unchanged, since perturbing whole-body states easily breaks balance. Motion diversity is thus bounded by what the operator demonstrates, and physics-aware trajectory augmentation is a natural next step.

Second, our simulation fidelity depends on automatically generated assets, whose geometry and physical parameters may be inaccurate for visually complex objects, widening the sim-to-real gap on contact-rich tasks. Better asset reconstruction and physical-parameter calibration could help close this gap.

\acknowledgments{}
This work is supported by the National Key Research and Development Program of China (Grant No.2024YFE0210900), the National Natural Science Foundation of China (Grant No.62306242),  the Young Elite Scientists Sponsorship Program by CAST (Grant No. 2024QNRC001), and the Yangfan Project of the Shanghai (Grant No.23YF11462200).

\bibliography{main}

\clearpage
\appendix

\section{Material Density}
\label{appendix:material_density}
To enable contact-rich manipulation in simulation, each generated asset
is assigned a mass based on its mesh volume and a category-level
material density. Table~\ref{tab:material_density} lists the density
values used in our experiments.

\begin{table}[ht]
\centering
\setlength{\tabcolsep}{10pt}
\renewcommand{\arraystretch}{1.15}
\begin{tabular}{lcc}
\toprule
\textbf{Material} & \textbf{Density (kg/m\textsuperscript{3})} & \textbf{Example Object} \\
\midrule
Polypropylene (PP)  & 910   & Box \\
Polyurethane (foam) & 50    & Sponge \\
ABS                 & 1050  & Monitor \\
Wicker              & 200   & Basket \\
Wood                & 700   & Cup \\
\bottomrule
\end{tabular}
\vspace{1em}
\caption{Density values used for assigning physical mass to generated assets in simulation.}
\label{tab:material_density}
\end{table}

\section{Domain Randomization}
\label{appendix:domain_randomization}

We apply domain randomization during offline rendering. Table~\ref{tab:domain_randomization} lists the 
randomized parameters.

\begin{table}[ht]
\centering
\begin{tabular}{ll}
\toprule
\textbf{Parameter} & \textbf{Distribution} \\
\midrule
\multicolumn{2}{l}{\textbf{Background Materials}} \\
Wall diffuse texture                & $\mathcal{U}(\text{Concrete, Wood, Terrazzo, Metal})$ \\
Floor diffuse texture               & $\mathcal{U}(\text{Concrete, Wood, Terrazzo})$ \\
Table diffuse texture               & $\mathcal{U}(\text{Wood})$ \\
Roughness                           & $\mathcal{U}(0.1,\ 0.65)$ \\
Metallic constant                   & $\mathcal{U}(0.25,\ 1.0)$ \\
Texture rotation [deg]              & $\mathcal{U}(0,\ 45)$ \\
Texture translation                 & $\mathcal{U}(0.1,\ 1.0)$ \\
UVW projection              & $\mathcal{B}(0.9)$ \\
\midrule
\multicolumn{2}{l}{\textbf{Lighting}} \\
Dome light intensity                 & $\mathcal{U}(1000,\ 3000)$ \\
Dome light color temperature     & $\mathcal{U}(4500,\ 6500)$ \\
Dome light color (RGB)               & $\mathcal{U}(0.85, 1.0) \times \mathcal{U}(0.85, 1.0) \times \mathcal{U}(0.85, 1.0)$ \\
Indoor light intensity               & $\mathcal{U}(20000,\ 200000)$ \\
Indoor light color temperature   & $\mathcal{U}(4500,\ 6500)$ \\
Indoor light color (RGB)             & $\mathcal{U}(0.85, 1.0) \times \mathcal{U}(0.85, 1.0) \times \mathcal{U}(0.85, 1.0)$ \\
\midrule
\multicolumn{2}{l}{\textbf{Camera Extrinsics}} \\
Position offset $(x, y, z)$ [m] & $\mathcal{U}(-0.01, 0.01) \times \mathcal{U}(-0.01, 0.01) \times \mathcal{U}(-0.01, 0.01)$ \\
Rotation offset (roll, pitch, yaw) [deg]   & $\mathcal{U}(-1.5, 1.5) \times \mathcal{U}(-1.5, 1.5) \times \mathcal{U}(-1.5, 1.5)$ \\
\bottomrule
\end{tabular}

\vspace{1em}
\caption{\textbf{Domain randomization parameters used during offline rendering.}
$\mathcal{U}(a, b)$ denotes a uniform distribution over $[a, b]$, 
and $\mathcal{B}(p)$ denotes a Bernoulli distribution with success 
probability $p$.}
\label{tab:domain_randomization}
\end{table}

\section{Real-to-Sim Asset Generation Details}
\label{appendix:asset_generation}

\subsection{Prompt Template for Physical Attribute Estimation}
\label{appendix:prompt}

We query Qwen3-VL with the reference image and a category description 
of the object, using the following prompt:

\begin{quote}
\itshape
This is a 3D model of \{category\}. Estimate its real-world dimensions 
in centimeters (length $\times$ width $\times$ height). Consider typical 
sizes of this object category. Output JSON: \{``length\_cm'': X, 
``width\_cm'': Y, ``height\_cm'': Z, ``material'': ``''\}
\end{quote}

The model's output is parsed as JSON to populate the physical dimensions 
and material category of the corresponding 3D asset.

\subsection{Dimension Accuracy Evaluation}
\label{appendix:dimension_eval}

To assess the reliability of Qwen3-VL's physical dimension estimation, 
we compare the predicted dimensions against ground-truth measurements 
obtained with a caliper on 5 real-world objects. 
Table~\ref{tab:dimension_eval} reports the per-object predicted and measured 
dimensions, along with the relative error.

\begin{table}[ht]
\centering

\begin{tabular}{lccc}
\toprule
\textbf{Object} & \textbf{Predicted (cm)} & \textbf{Measured (cm)} & \textbf{Avg. Error (cm)} \\
& $(L \times W \times H)$ & $(L \times W \times H)$ &  \\
\midrule
Box          & $26 \times 19 \times 11$  & $22 \times 21 \times 10$  & 2.3 \\
Sponge         & $24 \times 11 \times 6$ & $20 \times 9.5 \times 4.5$ & 2.3 \\
Monitor        & $61 \times 21 \times 46$ & $61 \times 23 \times 45$ & 1.0 \\
Basket       & $26 \times 24 \times 18$  & $30 \times 25 \times 22$  & 3.0 \\
Cup       & $7 \times 7 \times 12$  & $7 \times 7 \times 11$  & 0.3 \\

\bottomrule
\end{tabular}
\vspace{1em}
\caption{Comparison between Qwen3-VL predicted dimensions and real-world 
measurements. }
\label{tab:dimension_eval}
\end{table}

\section{Ablation on Curriculum-based Rollout}
\label{appendix:rollout_ablation}

To validate the necessity of the curriculum-based rollout mechanism, 
we compare two training variants:
\begin{itemize}
    \item \textbf{w/o Rollout}: the planner is trained exclusively on 
    ground-truth history.
    \item \textbf{w/ Rollout}: the curriculum-based rollout 
    described in Sec.~\ref{sec:training_recipe}.
\end{itemize}

We evaluate both variants on four manipulation tasks and report success 
rate per task as well as the overall average. Results are summarized in 
Table~\ref{tab:rollout_ablation}.

\begin{table}[ht]
\centering
\begin{tabular}{lcccc}
\toprule
\textbf{Variant} & \makecell{\textbf{Place Cup}\\\textbf{in Box}} & \makecell{\textbf{Wipe}\\\textbf{Monitor}} & \makecell{\textbf{Lift Basket and}\\\textbf{Place Cup}} & \makecell{\textbf{Kneel and Wipe}\\\textbf{Under Table}} \\
\midrule
w/o Rollout & 2/10 & 1/10 & 0/10 & 0/10 \\
w/ Rollout & \textbf{8/10} & \textbf{8/10} & \textbf{7/10} & \textbf{10/10} \\
\bottomrule
\end{tabular}
\vspace{1em}
\caption{\textbf{Ablation on the curriculum-based rollout mechanism.} 
Training without rollout leads to compounding errors over long horizons, 
resulting in consistently lower success rates across all tasks.}
\label{tab:rollout_ablation}
\end{table}

\end{document}